\documentclass{article}
\usepackage{spconf,amsmath,graphicx}

\usepackage{times}
\usepackage{graphicx}
\usepackage{amsmath,amssymb} 
\usepackage{epstopdf}
\usepackage{color}

\usepackage{cite}
\usepackage{subfigure}
\usepackage{multirow}
\usepackage{diagbox}
\usepackage{rotating}
\usepackage{booktabs}
\usepackage{colortbl}
\usepackage{overpic}
\usepackage[table]{xcolor}
\usepackage{textcomp}
\usepackage{contour}
\usepackage{textcomp}

\graphicspath{{./Figs/}}
\usepackage{float}
\usepackage[skip=5pt]{caption}

\definecolor{mygreen}{RGB}{0,150,0}
\definecolor{myred}{RGB}{200,0,0}

\renewcommand{\tabcolsep}{.5mm}

\graphicspath{{./Imgs/}}

\title{A fixation-based \textsc{360$^\circ$} benchmark dataset for salient object detection}
\name{Yi Zhang$^\dagger$, Lu Zhang$^\dagger$, Wassim Hamidouche$^\dagger$, Olivier Deforges$^\dagger$}
\address{$^\dagger$~IETR INSA Rennes, France}
\begin{document}
%
\maketitle
\begin{abstract}
Fixation prediction (FP) in panoramic contents has been widely investigated along with the booming trend of virtual reality (VR) applications. However, another issue within the field of visual saliency, salient object detection (SOD), has been seldom explored in 360$^\circ$ (or omnidirectional) images due to the lack of datasets representative of real scenes with pixel-level annotations. Toward this end, we collect 107 equirectangular panoramas with challenging scenes and multiple object classes. Based on the consistency between FP and explicit saliency judgements, we further manually annotate 1,165 salient objects over the collected images with precise masks under the guidance of real human eye fixation maps. Six state-of-the-art SOD models are then benchmarked on the proposed fixation-based 360$^\circ$ image dataset (\textbf{F-360iSOD}), by applying a multiple cubic projection-based fine-tuning method. Experimental results show a limitation of the current methods when used for SOD in panoramic images, which indicates the proposed dataset is challenging. Key issues for 360$^\circ$ SOD is also discussed. The proposed dataset is available at \it{https://github.com/PanoAsh/F-360iSOD}.

\end{abstract}
\section{Introduction}
\label{sec:intro}
The panoramic image, or 360$^\circ$ (omnidirectional) image, which captures the content on the whole 360$\times$180$^\circ$ viewing range surrounding a viewer, plays an import role in virtual reality (VR) applications and distinguishes itself from traditional 2-dimensional (2D) image which covers only one specific plane. Recently, commercial Head-Mounted Displays (HMDs) are developed to provide observers an immersive even interactive experience by allowing them to freely rotate their head and thus focusing on desired scenes and objects. Considering the fact that some salient parts of the  360$^\circ$ image attract more human attentions than others \cite{stfvr}, visual saliency prediction in panoramas becomes one of the focused issues within the field of computer vision and is considered as a key to study human observation behavior in virtual environments. The fixation prediction (FP) and salient object detection (SOD) are both closely related to the concept of visual saliency. Thanks to the accessibility of HMDs and eye

\begin{figure}[H]
  \centering
  \begin{overpic}[width=1\columnwidth]{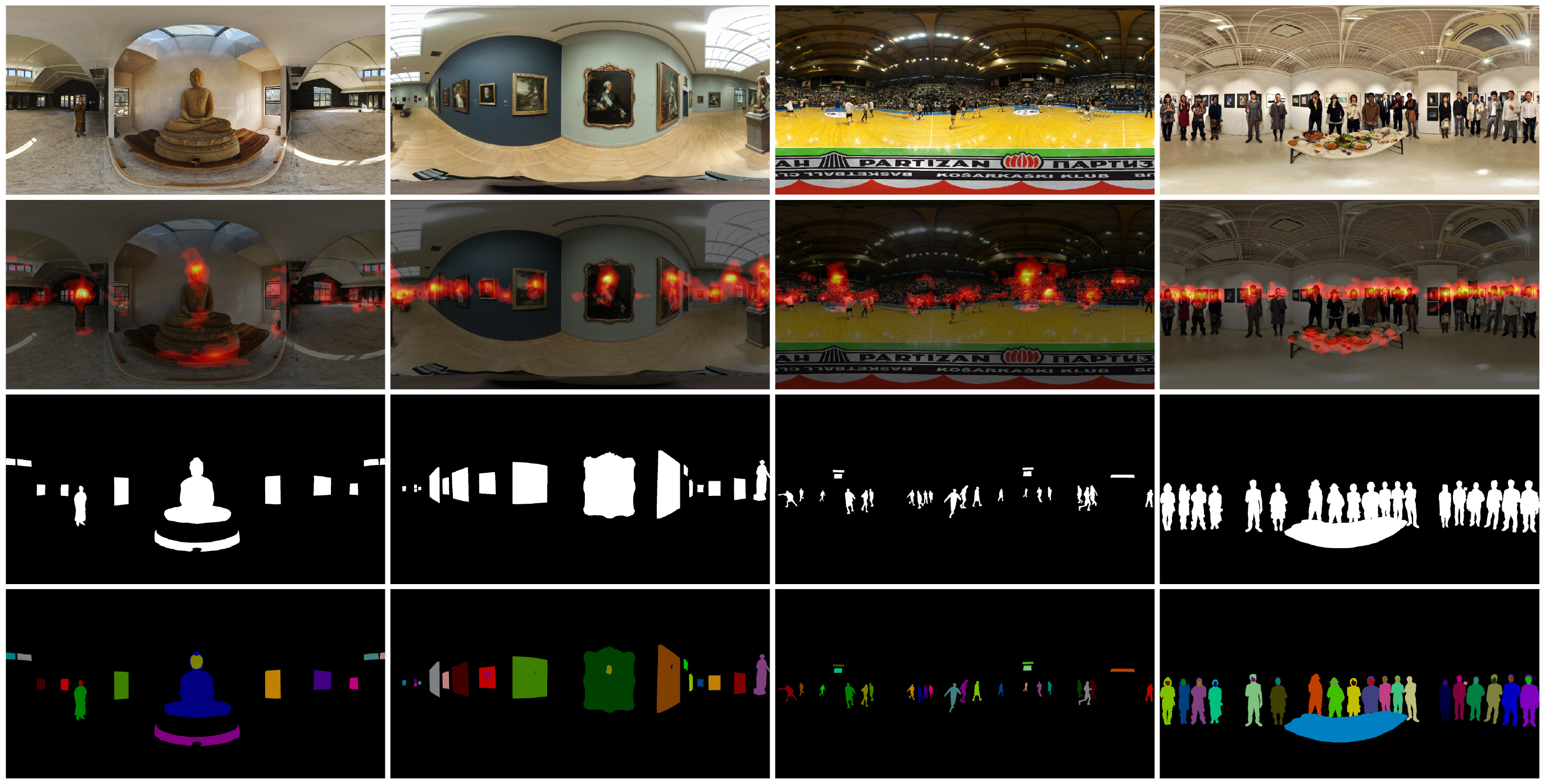}
  \end{overpic}
  \caption{Representative samples of the proposed fixation-based 360$^\circ$ image dataset (\textbf{F-360iSOD}). The first row: four panoramic images presented in the equirectangular format; the second row: images overlayed with thresholded fixation maps; the third row: object-level ground-truths; the fourth row: instance-level ground-truths.}\label{fig:1}
\end{figure}

\vspace{-10pt}
\noindent 
trackers, image \cite{salient360img} and video (e.g., \cite{vqaov, vrscene, saliency360video}) datasets have been constructed for the deep learning-based FP in panoramic content. However, to the best of our knowledge, \cite{360SOD} is the only study for SOD in 360$^\circ$ scenarios, which does not use the fixations as a guidance for the salient object annotation. 

As shown in Fig. 1, 360$^\circ$ images tend to have richer scenes and much more foreground objects compared to flat-2D images from traditional SOD datasets (e.g., \cite{ECSSD,SOD,PASCALS,HKUIS,DUTS,DUTO}). Therefore, it is more challenging to differentiate the salient objects from the non-salient ones in panoramas. Preserving 360$^\circ$ images with a few obvious foreground objects while discarding those ambiguous ones may bring selection bias to the dataset, thus being inefficient for exploring the real human attention behavior as viewing panoramic content. Based on the strong correlation between FP and explicit human judgements \cite{JUDDA}, and the successfully established fixation-based 2D SOD datasets \cite{JUDDA,LijiaVOS,DAVSOD}, we argue that the salient objects in panoramas can also be manually annotated with the assistance of fixations, thus representing the real-world daily scenes. The main contributions of this paper are: \textbf{1)} a fixation-based 360$^\circ$ image dataset (\textbf{F-360iSOD}) with both the object-/instance-level pixel-wise annotations is proposed; \textbf{2)} six newly proposed state-of-the-art 2D SOD models \cite{BASNet,EGNet,poolnet,CPD,SCRN,GCPANet} are benchmarked by five widely used SOD metrics \cite{Fmeasure,wfmeasure,Fan2017Smeasure,MAE,Fan2018Enhanced} on the proposed dataset in a cross-testing manner, with a multiple cubic projection-based fine-tuning strategy \cite{salgan360}; \textbf{3)} key issues for 360$^\circ$ SOD are discussed.

\section{Related work}
\label{sec:related work}

\subsection{2D SOD Datasets}
\label{ssec:2d sod datasets}
ECSSD~\cite{ECSSD}, SOD \cite{SOD}, PASCAL-S \cite{PASCALS}, HKU-IS \cite{HKUIS}, DUTS \cite{DUTS} and DUT-OMRON \cite{DUTO} are the six most widely used image datasets to benchmark the deep learning-based 2D SOD models, these datasets all provide improved pixel-wise object-level annotations and high challenges. It is worth mentioning that many deep learning-based SOD models are trained on the training set of the DUTS since 2017. The DUTS is so far the largest image dataset for SOD which contains 10,553 images for training and a testing set of 5,019 images. SOC \cite{fan2018SOC} is a more recently proposed SOD dataset with 6,000 images and about 80 category labels. The dataset includes 3,000 images without salient objects (or with only the presence of background objects) and both the precise object-/instance-level ground-truths. Further research \cite{Zhang2020UCNet} also emphasizes the great importance of image depth (D) information by proposing RGB-D-based SOD models. Besides, similar to JUDD-A \cite{JUDDA} (image SOD), there are two video-based SOD datasets \cite{LijiaVOS,DAVSOD} which also apply fixations to aid the manual annotation of salient objects, thus providing new ideas for the future SOD dataset construction.

\vspace{-10pt}
\subsection{SOD Models for 2D Images}
\label{ssec:2d sod models}
In recent years, fully convolutional network (FCN)-based models dominate the field of SOD. The FCN-based architecture differentiates itself from other deep learning methods by giving saliency maps as outputs, rather than classification scores, thus managing to predict saliency maps in a single feed-forward process by benefiting from an end-to-end learning. EGNet \cite{EGNet} is one of the recently proposed top-performanced state-of-the-art models. The method is motivated by the idea that simultaneously learning the salient edge and object information can help improving performance of SOD models. It models these two complementary information with an independent network outside the VGG-based backbone. SCRN \cite{SCRN} is another newly proposed SOD model that considers the edge information. It also implements the SOD and salient edge detection in a synchronous way, by stacking several so-called cross refinement units in an end-to-end manner. BASNet \cite{BASNet} proposes residual refinement module and hybrid loss to refine the salient objects boundaries in predicted saliency maps. PoolNet \cite{poolnet} considers improving the feature extraction efficiency of multiple layers of current U shape architecture by adding two new modules, which are both designed based on simple pooling techniques. GCPANet \cite{GCPANet} is a more recently proposed method which brings improvements to the traditional bottom-up/top-down networks by proposing four new modules. CPD \cite{CPD} modifies the traditional encoder-decoder framework to directly refine high-level features by generated saliency maps, without the consideration of low-level features. The idea here is different from PoolNet and GCPANet, which try to integrate both the low-/high-level features. Due to the limited space, we will not include all the SOD models in this section (see recent benchmark studies \cite{fan2018SOC} for more).

\vspace{-10pt}
\subsection{Panoramic Datasets}
\label{ssec:panoramic datasets}
As topic-related, there are two types of panoramic datasets focusing on head movement (HM) prediction and FP, respectively. Datasets such as 360-VHMD \cite{360VHMD}, VR-VQA48 \cite{vrvqa48} and PVS-HM \cite{pvshm} contain only head tracking data, while Salient!360 \cite{salient360img}, Stanford360 \cite{stfvr}, VQA-OV \cite{vqaov}, VR-scene \cite{vrscene} and 360-Saliency \cite{saliency360video} provide ground-truth eye fixations. Besides, 360-SOD \cite{360SOD} is a newly proposed omnidirectional image dataset for SOD. However, the salient objects are labeled with pure human judgements, rather than fixation-based guidance. Besides, the dataset does not provide instance-level ground-truths or object class labels.

\section{A New Dataset for panoramic SOD}
\label{sec:dataset}

\vspace{-10pt}
\begin{figure}[H]
  \centering
  \begin{overpic}[width=0.9\columnwidth]{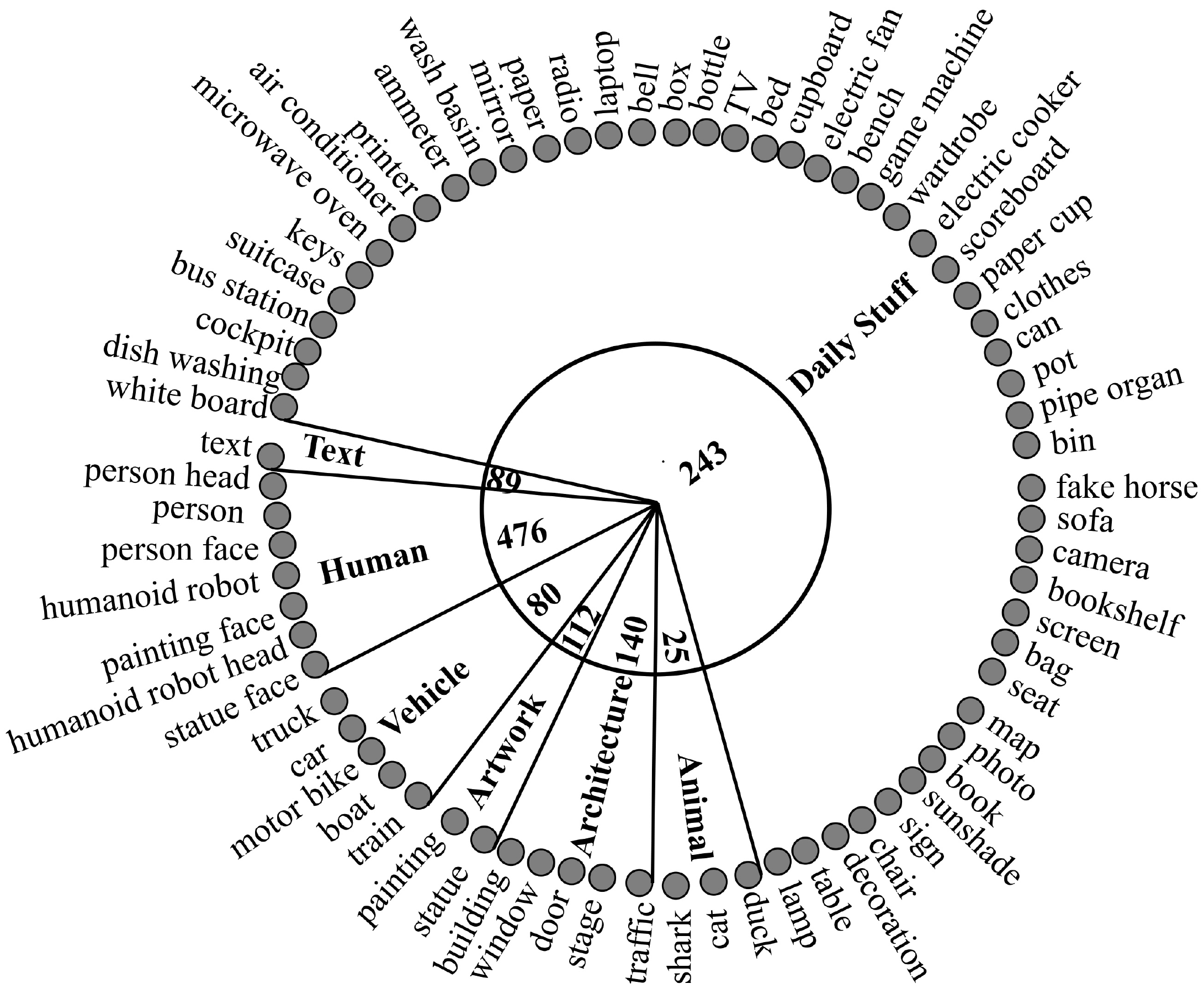}
  \end{overpic}
  \caption{Statistics of the proposed dataset (\textbf{F-360iSOD}).}\label{fig:2}
\end{figure}

\vspace{-10pt}
In this section, we present a new fixation-based 360$^\circ$ image dataset, called \textbf{F-360iSOD}, which contains 107 (52 indoor/55 outdoor) panoramic images with challenging real-world daily scenes, 1,165 salient objects (from 72 object classes) manually labeled with precise object-/instance-level masks.

\vspace{-10pt}
\subsection{Image Collection}
\label{ssec:image collection}
The \textbf{F-360iSOD} is a small-scale 360$^\circ$ dataset with totally 107 panoramic images collected from Stanford360 \cite{stfvr} and Salient!360 \cite{salient360img} which contain 85 and 22 equirectangular images, respectively. To the best of our knowledge, the Stanford360 and Salient!360 are the only panoramic image datasets that provide raw eye fixation data. All the images of the proposed \textbf{F-360iSOD} are presented in equirectangular format with a size of 2048$\times$1024 for convenient processing.

\vspace{-10pt}
\subsection{Salient Object Annotation}
\label{ssec:salient object annotation}
Inspired by the 2D SOD datasets \cite{JUDDA,LijiaVOS,DAVSOD} where fixation data are used to aid the salient object annotation, an expert is asked to manually annotate (by tracing boundaries) the salient objects with both the object-/instance-level masks on the collected equirectangular images, under the guidance of fixation maps convoluted by a Gaussian with a standard deviation empirically set to 3.34$^\circ$ of visual angle \cite{salient360img} (note that each of the Gaussian-smoothed fixation maps is thresholded with an adaptive saliency value to keep the top one-10th of each self before shown to the annotator). The whole annotation process has been repeated three times to pass the quality check implemented by two other experts, for the final ground-truths. Besides, nine images without any salient object annotations are kept in \textbf{F-360iSOD}, to avoid the common bias of 2D SOD datasets (as mentioned in \cite{YiSurvey}), brought by an assumption that there is at least one salient object in each of the image.

\vspace{-10pt}
\subsection{Dataset Statistics}
\label{ssec:dataset statistics}
In \textbf{F-360iSOD}, each of the salient object belongs to one specific class. Generally, there are 1,165 salient objects from 72 categories, thus reflecting 7 aspects (human, text, vehicle, architecture, artwork, animal and daily stuff) of the real-world common scenes (Fig. 2). The {\it person} category occupies the largest proportion with a number of instances of 386; other relative large object classes include {\it painting}, {\it text}, {\it building}, {\it person face} and {\it car}, with a number of instances of 92, 89, 86, 75 and 72, respectively.

\section{Experimental results and discussions}
\label{sec:experiments and discussions}

\subsection{Dataset Split}
\label{ssec:dataset split}
The \textbf{F-360iSOD} consists of one training set and two testing set, which are denoted as F-360iSOD-train, F-360iSOD-testA and F-360iSOD-testB, respectively. The F-360iSOD-train contains 68 equirectangular images from the Salient!360 \cite{salient360img}, while the F-360iSOD-testA collects the remaining 17 (85 in total). Besides, the F-360iSOD-testB is established to enable the cross-testing for SOD models, with 22 images from another panoramic image dataset (Stanford360 \cite{stfvr}).

\vspace{-10pt}
\subsection{Projection Methods}
\label{ssec:projection methods}
By wearing HMDs, people are able to freely rotate their head to make multiple viewports focusing on the attractive regions of the surrounding 360$^\circ$ content. Based on this prior knowledge, we apply cubemap projection (where a 360$^\circ$ image is projected into 6 rectangular patches) to process 68 panoramic images (from F-360iSOD-train) with multiple rotation angles (0$^\circ$, 30$^\circ$, 60$^\circ$ both horizontally and vertically \cite{salgan360}). Thus, we gain 54 (6$\times$3$\times$3) patches representative of multiple fields of view for each of the 360$^\circ$ image. 3672 (54$\times$68) 2D patches (256$\times$256) are therefore generated and used as inputs for the fine-tuning of 2D SOD models.

\vspace{-10pt}
\subsection{Evaluation Metrics}
\label{ssec:metrics}
To measure the agreement between manually labeled ground-truths and model predictions, we adopt five widely used SOD metrics: F-measure curves \cite{Fmeasure}, weighted $F_{\beta}$ measure (Fbw) \cite{wfmeasure}, mean absolute error (MAE) \cite{MAE}, structural measure (S-measure) \cite{Fan2017Smeasure} and enhanced-alignment measure (E-measure) curves \cite{Fan2018Enhanced}. Note that the $\beta^2$ is set to 0.3 in F-measure and Fbw, aiming to emphasize more on precision as suggested in \cite{Fmeasure}. F-measure, MAE and Fbw address the pixel-wise errors, while S-measure evaluates the structural similarity between predicted saliency maps and binary ground-truths :
\vspace{-5pt}
\begin{equation}\label{equ:2}
   S = \alpha{\times{S_{o}}}+(1-\alpha)\times{S_{r}},
   \vspace{-5pt}
\end{equation}
where $S_{o}$ and $S_{r}$ denote the object-/region-aware structure similarities, respectively; $\alpha$ is empirically set to 0.7 ($\alpha=0.5$ in 2D) to attach more importance on object structure, based on the observation that panoramic images are usually dominated by small salient objects distributed over the whole image (e.g., Fig. 1), rather than one or multiple spatially connected foreground objects located at the center of the image. E-measure is a more recently proposed SOD metric which combines both the pixel-/image-level information:
\vspace{-5pt}
\begin{equation}\label{equ:2}
   Q_{FM} = \frac{1}{W\times{H}}\Sigma_{i=1}^{W}\Sigma_{j=1}^{H}\phi_{FM}(x,y), 
   \vspace{-5pt}
\end{equation}
where the $\phi_{FM}$ means the enhanced alignment matrix; the $H$ and $M$ are the height and width of the foreground map.

\vspace{-5pt}
\subsection{Benchmarking Results}
\label{ssec:benchmarking results}

\vspace{-5pt}
\begin{figure}[H]
  \centering
  \begin{overpic}[width=1\columnwidth]{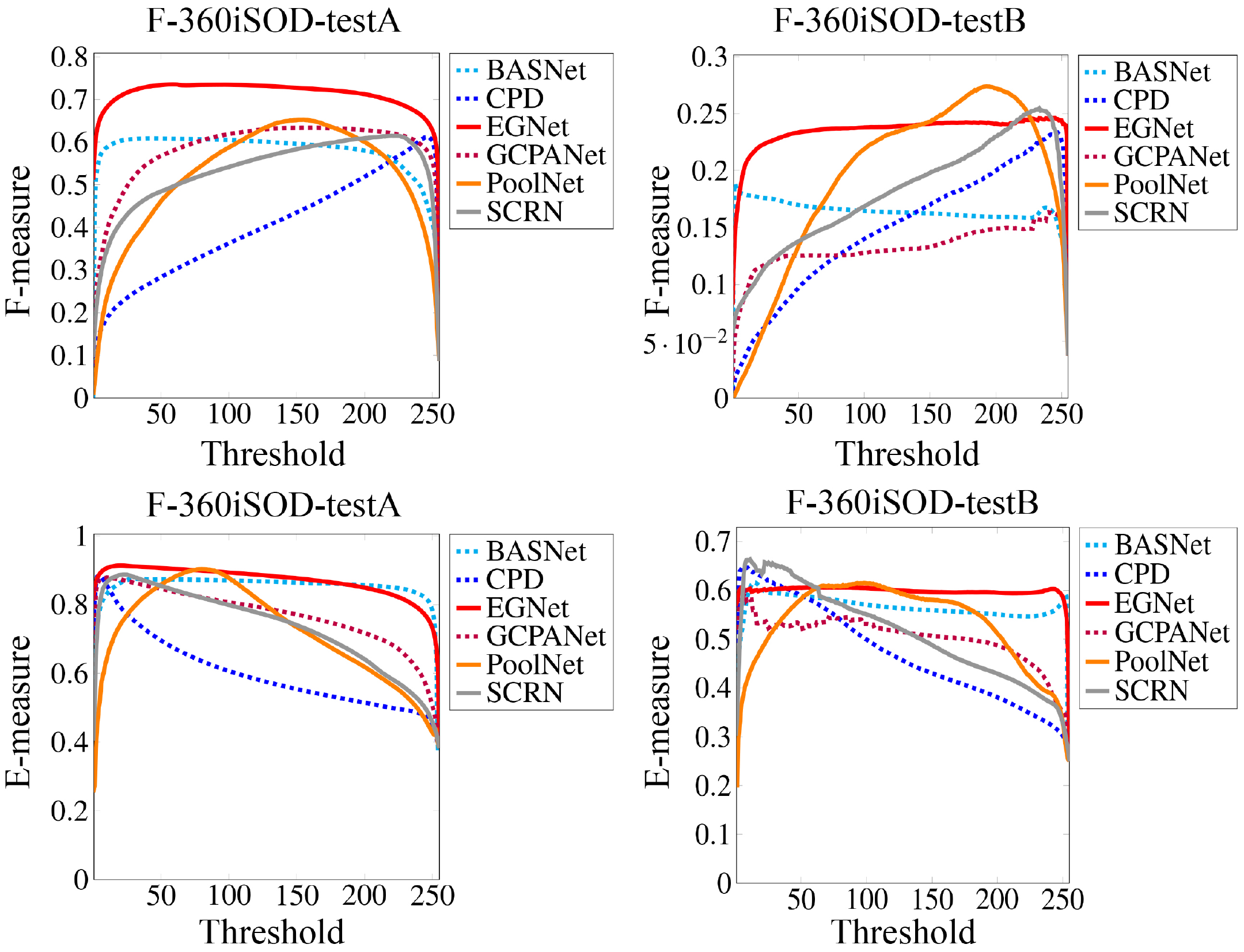}
  \end{overpic}
  \caption{F-measure curves and E-measure curves obtained by six state-of-the-art SOD models on the \textbf{F-360iSOD}.}\label{fig:3}
\end{figure}

\begin{figure*}[ht!]
  \centering
  \begin{overpic}[width=2\columnwidth]{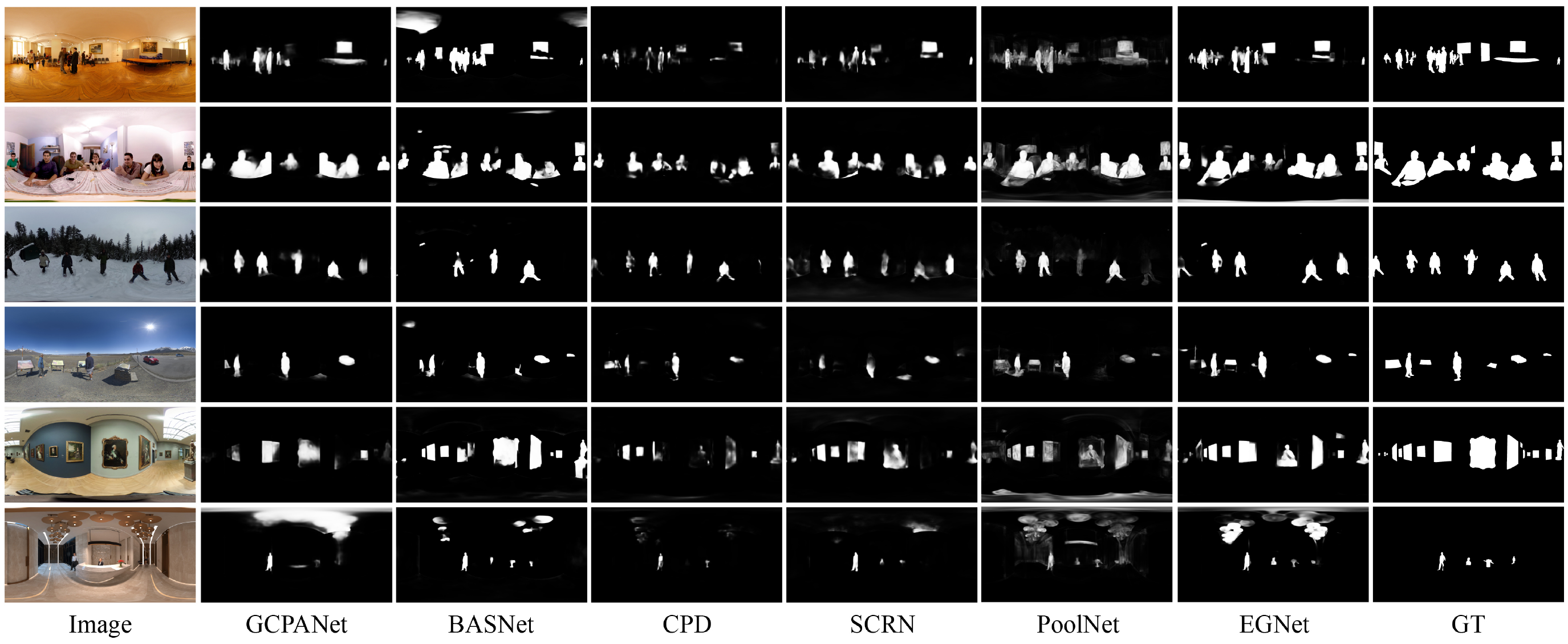}
  \end{overpic}
  \caption{A qualitative comparison between six state-of-the-art SOD models on \textbf{F-360iSOD}.}\label{fig:4}
  \vspace{-10pt}
\end{figure*}

\vspace{-10pt}
In our study, each of the SOD model is fine-tuned on the F-360iSOD-train with an initial learning rate of one-10th of the default, and a batch size of 1. The training process will stop as the S-measure value on the F-360iSOD-testA starts to go down. As a result, it takes about 20 epochs for BASNet \cite{BASNet}, EGNet \cite{EGNet}, CPD \cite{CPD} and SCRN \cite{SCRN} to converge, while 70 for PoolNet \cite{poolnet} and 15 for GCPANet \cite{GCPANet}. The quantitative and qualitative comparison between the six state-of-the-art 2D SOD models on both the F-360iSOD-testA/B are illustrated in Table. 1, Fig. 3 and Fig. 4, respectively. 

\begin{table}[H]
  \centering
  \renewcommand{\arraystretch}{0.5}
  \renewcommand{\tabcolsep}{0.65mm}
  \begin{tabular}{r||ccc|ccc}
   \toprule
  \multirow{2}{*}{Methods~~}  & \multicolumn{3}{c}{F-360iSOD-testA} & \multicolumn{3}{c}{F-360iSOD-testB}
  \\
  \cline{2-4} \cline{5-7} 
  \\
    &$F_{\beta}^{w}~\uparrow$ & $S~\uparrow$ & $MAE~\downarrow$ & $F_{\beta}^{w}~\uparrow$ & $S~\uparrow$& $MAE~\downarrow$\\
  \midrule
  \\ 
  
   SCRN \cite{SCRN}  & .551 & .809 &  \textcolor{blue}{.050}  & .124 &  \textcolor{blue}{.708} &  \textcolor{green}{.034 } \\\\
   
   BASNet \cite{BASNet} & \textcolor{blue}{.567} & \textcolor{blue}{.825}  &  \textcolor{green}{.046}  & .118 & .683 &  .048\\\\
   
   CPD \cite{CPD} &  .521 & .763 &  .052 &  \textcolor{blue}{.129} & .695 &   \textcolor{red}{.032}  \\\\
   
   PoolNet \cite{poolnet} &  .500 & \textcolor{green}{.834} &  .068 &  \textcolor{green}{.136} &  \textcolor{red}{.716} &  .058 \\\\
  
   GCPANet \cite{GCPANet} &  \textcolor{green}{.630} & .822 &  \textcolor{red}{.045}  &  .106 & .693  &  \textcolor{blue}{.039} \\\\
    
   EGNet \cite{EGNet} & \textcolor{red}{.715} & \textcolor{red}{.864} &  \textcolor{red}{.045} &  \textcolor{red}{.190} &  \textcolor{green}{.714} & .041 \\\\

  \bottomrule
  \end{tabular}
  \caption{A quantitative comparison between six state-of-the-art SOD models on \textbf{F-360iSOD}, where $F_{\beta}^{w}$ means Fbw, $S$ represents S-measure. Note that the top three results of each column are highlighted in \textcolor{red}{red}, \textcolor{green}{green} and \textcolor{blue}{blue}, respectively.}
\end{table}

\vspace{-30pt}
\subsection{Discussions}
\label{ssec:discussions}
\textbf{Features of 360$^\circ$ datasets.} All the benchmarking models are constrained to some extent on the proposed \textbf{F-360iSOD}, even though achieving high performances in 2D SOD \cite{BASNet,EGNet,poolnet,CPD,SCRN,GCPANet}. The limitation is mainly due to the challenges brought by the features of 360$^\circ$ dataset, such as equirectangular projection-induced distortions, small objects and clutter scenes, etc. 

\noindent
\textbf{Fixation-based complexity analysis.} Since the panoramic images tend to contain much more scenes and objects than 2D images, the ambiguity of saliency judgements in panoramas should also be considered, which can be quantified by inter observer congruency (IOC) \cite{IOCanalysis} and entropy based on fixation maps, which are re-smoothed with a Gaussian with a standard deviation of 1$^\circ$ visual angle to reflect human foveal size \cite{IOCanalysis}. As an image with high IOC and low entropy is usually considered to be simple, the F-360iSOD-testB should be easier to explore when compared with the F-360iSOD-testA (Fig. 5), from a perspective of human judgements.

\noindent
\textbf{Unseen object classes.} All benchmarking models fail on the 

\begin{figure}[H]
  \centering
  \begin{overpic}[width=0.9\columnwidth]{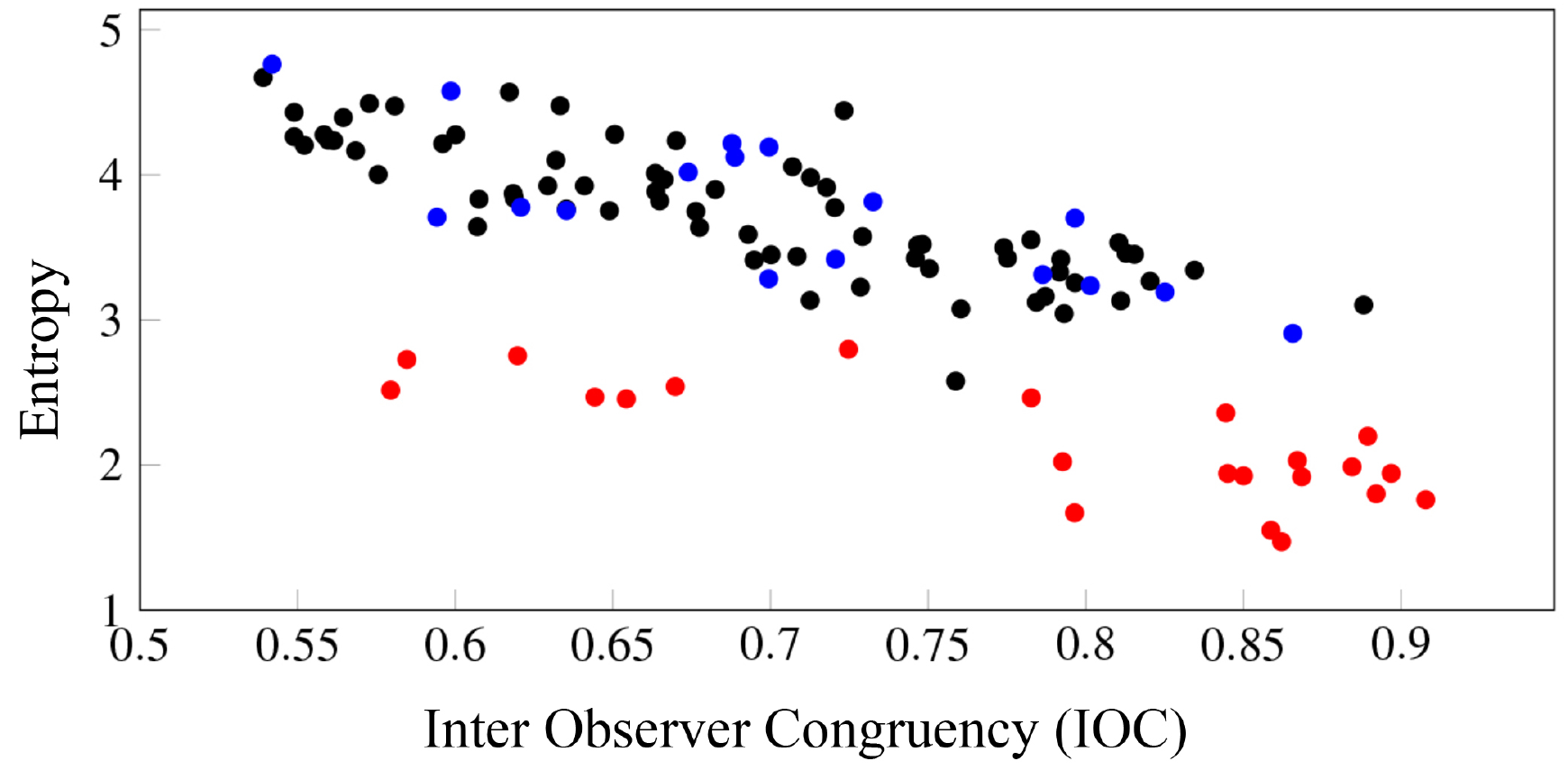}
  \end{overpic}
  \caption{A fixation-based complexity analysis of the proposed \textbf{F-360iSOD}. The F-360iSOD-train, F-360iSOD-testA/B are annotated in \textcolor{black}{black}, \textcolor{blue}{blue} and \textcolor{red}{red}, respectively.}\label{fig:5}
\end{figure}

\vspace{-10pt}
\noindent
F-360iSOD-testB, mainly due to the presence of unseen object classes in Stanford360 \cite{stfvr}, such as sharks, bells, robots, etc. People are capable of recognizing new object categories when provided with high-level descriptions. This strong generalization ability is still absent in current SOD models. 

\noindent
\textbf{Instance-level ground-truths.} To the best of our knowledge, the proposed \textbf{F-360iSOD} is the first 360$^\circ$ dataset that provides instance-level semantic labels for salient objects. Future SOD models are capable of recognizing the individual instances from multiple classes, which is crucial for practical applications, e.g., image captioning and scene understanding.

\vspace{-5pt}
\section{Conclusions}
\label{sec:conclusions}
In this paper, we propose a fixation-based 360$^\circ$ image dataset (\textbf{F-360iSOD}), with precisely annotated salient objects/instances from multiple classes representative of real-world daily scenes. Six recently proposed top-performanced SOD methods are fine-tuned and tested on the \textbf{F-360iSOD}. Results show a limit of current 2D models when directly applied to the SOD in panoramas. We believe the \textbf{F-360iSOD} can be used as one of the basic panoramic datasets and thus supporting future 360$^\circ$ SOD model development.

\bibliographystyle{f360isod}
{\ninept
\bibliography{f360isod}}

\begin{thebibliography}{10}

\bibitem{stfvr}
Vincent Sitzmann, Ana Serrano, Amy Pavel, Maneesh Agrawala, Diego Gutierrez,
  Belen Masia, and Gordon Wetzstein,
\newblock ``Saliency in vr: How do people explore virtual environments?,''
\newblock {\em IEEE TVCG}, vol. 24, no. 4, pp. 1633--1642, 2018.

\bibitem{salient360img}
Yashas Rai, Jes{\'u}s Guti{\'e}rrez, and Patrick Le~Callet,
\newblock ``A dataset of head and eye movements for 360 degree images,''
\newblock in {\em MMSys}. ACM, 2017, pp. 205--210.

\bibitem{vqaov}
Chen Li, Mai Xu, Xinzhe Du, and Zulin Wang,
\newblock ``Bridge the gap between vqa and human behavior on omnidirectional
  video: A large-scale dataset and a deep learning model,''
\newblock in {\em ACM MM}, 2018, pp. 932--940.

\bibitem{vrscene}
Yanyu Xu, Yanbing Dong, Junru Wu, Zhengzhong Sun, Zhiru Shi, Jingyi Yu, and
  Shenghua Gao,
\newblock ``Gaze prediction in dynamic 360 immersive videos,''
\newblock in {\em IEEE CVPR}, 2018, pp. 5333--5342.

\bibitem{saliency360video}
Ziheng Zhang, Yanyu Xu, Jingyi Yu, and Shenghua Gao,
\newblock ``Saliency detection in 360 videos,''
\newblock in {\em ECCV}, 2018, pp. 488--503.

\bibitem{360SOD}
Jia Li, Jinming Su, Changqun Xia, and Yonghong Tian,
\newblock ``Distortion-adaptive salient object detection in 360°
  omnidirectional images,''
\newblock {\em IEEE JSTSP}, 2019.

\bibitem{ECSSD}
Qiong Yan, Li~Xu, Jianping Shi, and Jiaya Jia,
\newblock ``Hierarchical saliency detection,''
\newblock in {\em IEEE CVPR}, 2013, pp. 1155--1162.

\bibitem{SOD}
Vida Movahedi and James~H Elder,
\newblock ``Design and perceptual validation of performance measures for
  salient object segmentation,''
\newblock in {\em CVPRw}. IEEE, 2010, pp. 49--56.

\bibitem{PASCALS}
Yin Li, Xiaodi Hou, Christof Koch, James~M Rehg, and Alan~L Yuille,
\newblock ``The secrets of salient object segmentation,''
\newblock in {\em IEEE CVPR}, 2014, pp. 280--287.

\bibitem{HKUIS}
Guanbin Li and Yizhou Yu,
\newblock ``Visual saliency based on multiscale deep features,''
\newblock in {\em IEEE CVPR}, 2015, pp. 5455--5463.

\bibitem{DUTS}
Lijun Wang, Huchuan Lu, Yifan Wang, Mengyang Feng, Dong Wang, Baocai Yin, and
  Xiang Ruan,
\newblock ``Learning to detect salient objects with image-level supervision,''
\newblock in {\em IEEE CVPR}, 2017, pp. 136--145.

\bibitem{DUTO}
Chuan Yang, Lihe Zhang, Huchuan Lu, Xiang Ruan, and Ming-Hsuan Yang,
\newblock ``Saliency detection via graph-based manifold ranking,''
\newblock in {\em IEEE CVPR}, 2013, pp. 3166--3173.

\bibitem{JUDDA}
Ali Borji,
\newblock ``What is a salient object? a dataset and a baseline model for
  salient object detection,''
\newblock {\em IEEE TIP}, vol. 24, no. 2, pp. 742--756, 2014.

\bibitem{LijiaVOS}
Jia Li, Changqun Xia, and Xiaowu Chen,
\newblock ``A benchmark dataset and saliency-guided stacked autoencoders for
  video-based salient object detection,''
\newblock {\em IEEE TIP}, vol. 27, no. 1, pp. 349--364, 2017.

\bibitem{DAVSOD}
Deng-Ping Fan, Wenguan Wang, Ming-Ming Cheng, and Jianbing Shen,
\newblock ``Shifting more attention to video salient object detection,''
\newblock in {\em IEEE CVPR}, 2019, pp. 8554--8564.

\bibitem{BASNet}
Xuebin Qin, Zichen Zhang, Chenyang Huang, Chao Gao, Masood Dehghan, and Martin
  Jagersand,
\newblock ``Basnet: Boundary-aware salient object detection,''
\newblock in {\em IEEE CVPR}, 2019, pp. 7479--7489.

\bibitem{EGNet}
Jia-Xing Zhao, Jiang-Jiang Liu, Deng-Ping Fan, Yang Cao, Jufeng Yang, and
  Ming-Ming Cheng,
\newblock ``Egnet: Edge guidance network for salient object detection,''
\newblock in {\em IEEE ICCV}, 2019, pp. 8779--8788.

\bibitem{poolnet}
Jiang-Jiang Liu, Qibin Hou, Ming-Ming Cheng, Jiashi Feng, and Jianmin Jiang,
\newblock ``A simple pooling-based design for real-time salient object
  detection,''
\newblock in {\em IEEE CVPR}, 2019.

\bibitem{CPD}
Zhe Wu, Li~Su, and Qingming Huang,
\newblock ``Cascaded partial decoder for fast and accurate salient object
  detection,''
\newblock in {\em IEEE CVPR}, 2019, pp. 3907--3916.

\bibitem{SCRN}
Zhe Wu, Li~Su, and Qingming Huang,
\newblock ``Stacked cross refinement network for edge-aware salient object
  detection,''
\newblock in {\em IEEE ICCV}, 2019, pp. 7264--7273.

\bibitem{GCPANet}
Chen Zuyao, Xu~Qianqian, Cong Runmin, and Huang Qingming,
\newblock ``Global context-aware progressive aggregation network for salient
  object detection,''
\newblock in {\em AAAI}, 2020.

\bibitem{Fmeasure}
Radhakrishna Achanta, Sheila Hemami, Francisco Estrada, and Sabine
  S{\"u}sstrunk,
\newblock ``Frequency-tuned salient region detection,''
\newblock in {\em IEEE CVPR}, 2009, pp. 1597--1604.

\bibitem{wfmeasure}
Ran Margolin, Lihi Zelnik-Manor, and Ayellet Tal,
\newblock ``How to evaluate foreground maps?,''
\newblock in {\em IEEE CVPR}, 2014, pp. 248--255.

\bibitem{Fan2017Smeasure}
Deng-Ping Fan, Ming-Ming Cheng, Yun Liu, Tao Li, and Ali Borji,
\newblock ``Structure-measure: A new way to evaluate foreground maps,''
\newblock in {\em IEEE ICCV}, 2017, pp. 4548--4557.

\bibitem{MAE}
Federico Perazzi, Philipp Kr{\"a}henb{\"u}hl, Yael Pritch, and Alexander
  Hornung,
\newblock ``Saliency filters: Contrast based filtering for salient region
  detection,''
\newblock in {\em IEEE CVPR}, 2012, pp. 733--740.

\bibitem{Fan2018Enhanced}
Deng-Ping Fan, Cheng Gong, Yang Cao, Bo~Ren, Ming-Ming Cheng, and Ali Borji,
\newblock ``Enhanced-alignment measure for binary foreground map evaluation,''
\newblock {\em IJCAI}, pp. 698--704, 2018.

\bibitem{salgan360}
Fang-Yi Chao, Lu~Zhang, Wassim Hamidouche, and Olivier Deforges,
\newblock ``Salgan360: visual saliency prediction on 360 degree images with
  generative adversarial networks,''
\newblock in {\em ICMEw}. IEEE, 2018, pp. 01--04.

\bibitem{fan2018SOC}
Deng-Ping Fan, Ming-Ming Cheng, Jiang-Jiang Liu, Shang-Hua Gao, Qibin Hou, and
  Ali Borji,
\newblock ``Salient objects in clutter: Bringing salient object detection to
  the foreground,''
\newblock in {\em ECCV}, 2018, pp. 186--202.

\bibitem{Zhang2020UCNet}
Jing Zhang, Deng-Ping Fan, Yuchao Dai, Saeed Anwar, Fatemeh Sadat~Saleh, Tong
  Zhang, and Nick Barnes,
\newblock ``Uc-net: Uncertainty inspired rgb-d saliency detection via
  conditional variational autoencoders,''
\newblock in {\em IEEE CVPR}, 2020.

\bibitem{360VHMD}
Xavier Corbillon, Francesca De~Simone, and Gwendal Simon,
\newblock ``360-degree video head movement dataset,''
\newblock in {\em MMSys}. ACM, 2017, pp. 199--204.

\bibitem{vrvqa48}
Mai Xu, Chen Li, Yufan Liu, Xin Deng, and Jiaxin Lu,
\newblock ``A subjective visual quality assessment method of panoramic
  videos,''
\newblock in {\em ICME}. IEEE, 2017, pp. 517--522.

\bibitem{pvshm}
Mai Xu, Yuhang Song, Jianyi Wang, MingLang Qiao, Liangyu Huo, and Zulin Wang,
\newblock ``Predicting head movement in panoramic video: A deep reinforcement
  learning approach,''
\newblock {\em IEEE TPAMI}, 2018.

\bibitem{YiSurvey}
Yi~Zhang, Lu~Zhang, Wassim Hamidouche, and Olivier Deforges,
\newblock ``Key issues for the construction of salient object datasets with
  large-scale annotation,''
\newblock in {\em IEEE MIPR}, 2020.

\bibitem{IOCanalysis}
Olivier Le~Meur, Thierry Baccino, and Aline Roumy,
\newblock ``Prediction of the inter-observer visual congruency (iovc) and
  application to image ranking,''
\newblock in {\em ACM MM}, 2011, pp. 373--382.

\end{thebibliography}

\end{document}